\documentclass[10pt,twocolumn,letterpaper]{article}

\usepackage{iccv}  
\definecolor{iccvblue}{rgb}{0.21,0.49,0.74}
\usepackage[pagebackref,breaklinks,colorlinks,allcolors=iccvblue]{hyperref}
\usepackage{graphicx}
\usepackage{tabularx}
\usepackage{soul}
\usepackage{multirow}
\usepackage{amsmath,amsfonts}
\usepackage{amsthm}
\usepackage{pifont}
\usepackage{mathrsfs}
\usepackage{color}
\usepackage{soul}
\usepackage{colortbl}
\usepackage{tcolorbox}
\usepackage{textcomp}%
\usepackage{manyfoot}%
\usepackage{booktabs}%
\usepackage{algorithm}%
\usepackage{algorithmicx}%
\usepackage{algpseudocode}%
\usepackage{listings}%
\usepackage{float}
\usepackage{makecell}
\usepackage{ulem}
\usepackage{tcolorbox}
\usepackage{CJKutf8}
\tcbuselibrary{breakable}
\usepackage{orcidlink}

\def\paperID{*****} 
\def\confName{ICCV}
\def\confYear{2025}

\title{Ask2Loc: Learning to Locate Instructional Visual Answers by Asking Questions}

\makeatletter
\def\@fnsymbol#1{\ensuremath{\ifcase#1\or \or \ddagger\or
\mathsection\or \mathparagraph\or \|\or **\or \dagger\dagger
\or \ddagger\ddagger \else\@ctrerr\fi}}
\makeatother

\author{Chang Zong$^1$, Bin Li$^2$, Shoujun Zhou$^2$, Jian Wan$^3$, Lei Zhang$^1{^\dagger}$ \thanks{$^\dagger$Corresponding author (leizhang@zust.edu.cn)} \\ $^1$Zhejiang University of Science and Technology, Hangzhou, China \\ $^2$Shenzhen Institutes of Advanced Technology, Chinese Academy of Sciences, Shenzhen, China \\ $^3$Zhejiang University of Water Resources and Electric Power, Hangzhou, China}

\begin{document}
\newcommand{\best}{\cellcolor[HTML]{d5ecf5}}
\newcommand{\second}{\cellcolor[HTML]{4ffbdf}}
\newcommand{\impact}{\cellcolor[HTML]{f9f871}}
\newcommand{\improve}{\cellcolor[HTML]{fefedf}}
\definecolor{blue}{HTML}{d5ecf5}
\definecolor{green}{HTML}{4ffbdf}
\definecolor{gray}{HTML}{dcdcdc}
\definecolor{darkblue}{HTML}{0081cf}
\definecolor{red}{HTML}{c34a36}
\definecolor{yellow}{HTML}{fefedf}

\definecolor{blue2}{HTML}{F3C5FF}
\definecolor{darkblue2}{HTML}{845EC2}
\definecolor{green2}{HTML}{C4FCEF}
\definecolor{darkgreen2}{HTML}{00C2A8}

\newtcolorbox{systemprompt}{
    colback=blue2, 
    colframe=darkblue2,
    fonttitle=\bfseries,
    title=System Prompt, 
    breakable,          
    before upper={\parindent 0pt}
}

\newtcolorbox{userprompt}{
    colback=green2,
    colframe=darkgreen2,
    fonttitle=\bfseries,
    title=User Input,
    breakable,
    before upper={\parindent 0pt}
}

\maketitle
\begin{abstract}
Locating specific segments within an instructional video is an efficient way to acquire guiding knowledge. Generally, the task of obtaining video segments for both verbal explanations and visual demonstrations is known as visual answer localization (VAL). However, users often need multiple interactions to obtain answers that align with their expectations when using the system. During these interactions, humans deepen their understanding of the video content by asking themselves questions, thereby accurately identifying the location. Therefore, we propose a new task, named \textbf{In-VAL}, to simulate the multiple interactions between humans and videos in the procedure of obtaining visual answers. The In-VAL task requires interactively addressing several semantic gap issues, including 1) the ambiguity of user intent in the input questions, 2) the incompleteness of language in video subtitles, and 3) the fragmentation of content in video segments. To address these issues, we propose \textbf{Ask2Loc}, a framework for resolving In-VAL by asking questions. It includes three key modules: 1) a chatting module to refine initial questions and uncover clear intentions, 2) a rewriting module to generate fluent language and create complete descriptions, and 3) a searching module to broaden local context and provide integrated content. We conduct extensive experiments on three reconstructed In-VAL datasets. Compared to traditional end-to-end and two-stage methods, our proposed Ask2Loc can improve performance by up to 14.91 (mIoU) on the In-VAL task. Our code and datasets can be accessed at https://github.com/changzong/Ask2Loc.
\end{abstract}    

\section{Introduction}
\begin{figure}[t]
  \centering
  \includegraphics[width=.45\textwidth]{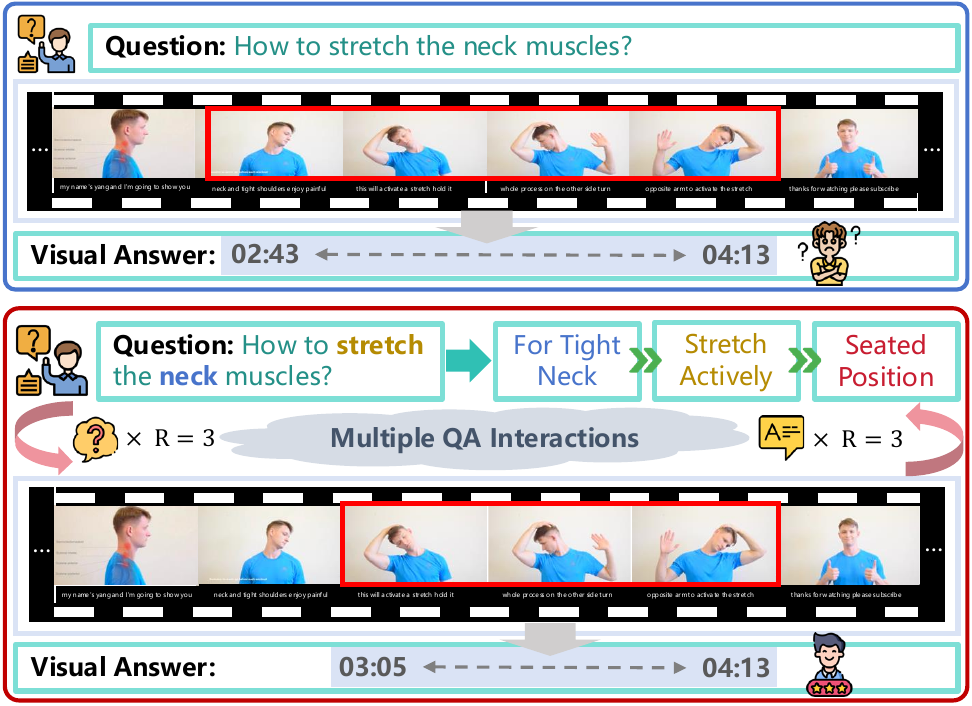}
  \caption{Existing VAL tasks derive visual answers directly from input questions, neglecting the natural interactions between the user and the video (above). In contrast, we propose an In-VAL task, which gradually acquires domain knowledge and clarifies user intent through multiple interactions to assist in answer localization (below).}
  \label{motivation}
\end{figure}
\begin{figure*}[ht]
  \centering
  \includegraphics[width=.9\textwidth]{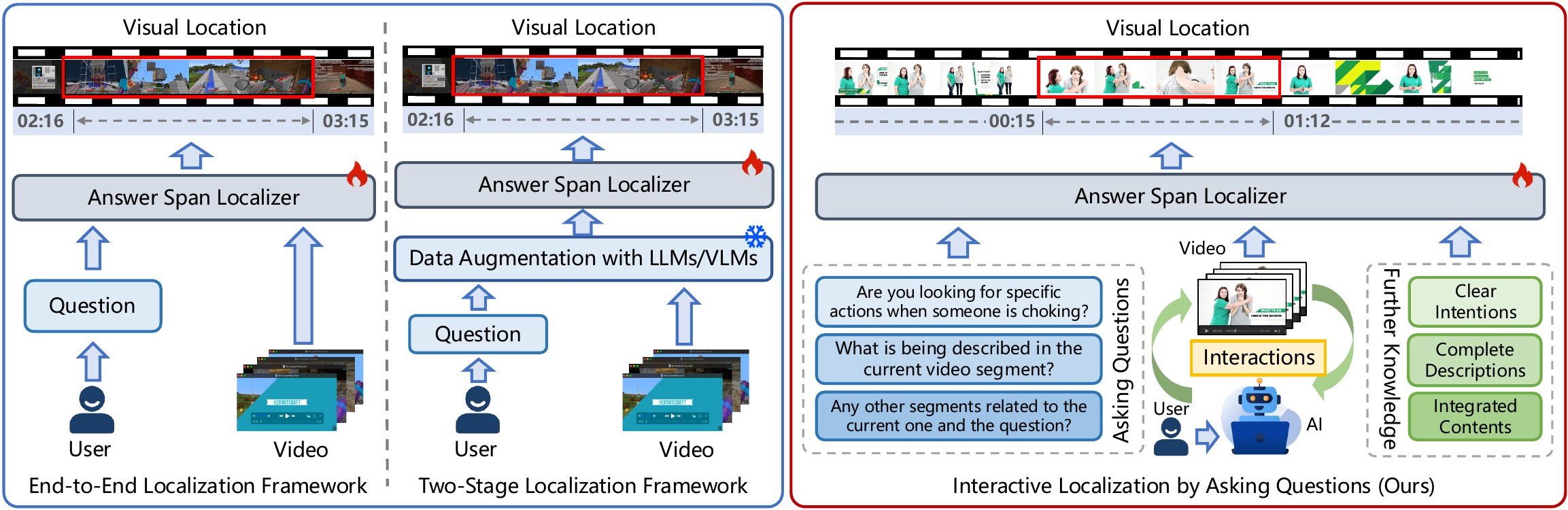}
  \caption{Apart from existing end-to-end and two-stage localization frameworks (left), our proposed framework (right) better simulates the interactions between humans and video content by leveraging an AI system. By asking multiple questions, the system can progressively acquire further understandings for solving the semantic issues, aiding in better localization.}
  \label{paradigm}
\end{figure*}

Instructional videos are designed to efficiently teach specific skills or procedures through step-by-step instructions ~\cite{zhukov2019cross,ghoddoosian2022hierarchical}. When users wish to be recommended the next action from an instructional video, they require both verbal explanations and visual demonstrations ~\cite{li2024towards}. In such a context, a challenging task has recently been proposed, named visual answer localization (VAL)~\cite{gupta2023dataset}. In this task, a frame span is expected to be located from a lengthy domain-specific (e.g., medical) video corresponding to a natural language question. However, in the VAL task, users may receive semantically correct but unexpected answers. Existing VAL tasks derive visual answers directly from input questions, neglecting the natural interactions between users and videos. When humans encounter a video in an unfamiliar domain, they typically need to interact with the video multiple times to obtain answers that align with their expectations. As shown in Figure \ref{motivation}, by leveraging the content of a video, users ask themselves a series of questions to gradually clarify their intents and acquire further knowledge, thereby achieving a better location. To address the neglect of this interaction in VAL, we propose a new task, named \textbf{In-VAL}, which aims to accomplish instructional visual answer localization by simulating the interactions between human users and videos.

With the advancement of multimodal techniques and large language models (LLMs), particularly the recent emergence of video language models, it is now possible to understand and reason with long videos ~\cite{cheng2023vindlu,maaz2023video,lin2023video}. However, to interactively identify video segments that can answer a question, the following semantic issues still need to be addressed: (1) Due to a lack of domain knowledge related to a video, the initial questions posed by users exhibit intent ambiguity. (2) The incompleteness of the verbal explanation in video segments hinders the understanding of the current content. (3) Constrained by the fragmentation of information in video segments, users still find it difficult to precisely locate the answer.

Existing studies for locating visual answers mainly follow an end-to-end or a two-stage framework. End-to-end methods focus on learning the deep fusion of visual features (frame images) and textual features (questions and subtitles) for predicting answer spans ~\cite{zhang2021natural,li2023learning,li2024towards,li2024fuzzy}. In contrast, two-stage methods alleviate inconsistencies in semantics and modalities by expanding the input question with an LLM or describing the visual content using a vision language model (VLM) ~\cite{zheng2024training,qu2024chatvtg}. However, these methods cannot be well adapted to the In-VAL task due to their inability to effectively simulate interactions between users and videos. For instance, users refine their input questions by quickly skimming through the full content of the video, thereby making their intents clearer. Additionally, users repeatedly view the preceding and succeeding segments of the current position to verify whether the content is relevant to their questions. By doing so, users can naturally align with the video content, thus supporting accurate localization.

Asking questions is a heuristic interactive way for query clarification and knowledge acquisition ~\cite{li2016learning, abdelghani2024gpt,sasson2025art} and has been widely used in QA systems ~\cite{kim2024prospector,kim2024qube,abdullin2024synthetic,10.1145/3616855.3635856} and video retrieval tasks ~\cite{10.1145/3503161.3548361}. Meanwhile, LLMs and pre-trained language models (PLMs) have demonstrated strong semantic understanding by learning from QA pairs ~\cite{jiang-etal-2024-instruction,singhal2025toward}. For instance, in dialogue systems, LLMs grasp multi-turn chatting content to capture the true intent of users, providing expected responses ~\cite{abdullin-etal-2023-synthetic,li2024hello,maharana-etal-2024-evaluating}. In reasoning tasks, LLMs achieve descriptive consistency by rewriting queries, facilitating subsequent analysis ~\cite{ma-etal-2023-query,sun2024r,liu2024query}. In domain QA tasks, PLMs improve answer quality by searching relevant documents as context ~\cite{asai2023self,edge2024local,sarmah2024hybridrag}. Inspired by these studies, we aim to develop an AI framework using LLMs and PLMs for chatting, rewriting, and searching to address the In-VAL task. By asking questions at these interactive steps, we can simulate how users gradually resolve semantic issues while locating visual answers (as described in Figure \ref{paradigm}).

To solve In-VAL, we propose \textbf{Ask2Loc}, an interactive visual answer localization framework via asking various questions using LLMs and PLMs. Specifically, we implement three interactive modules: (1) To tackle the intention ambiguity in input questions, we employ an LLM to simulate multi-turn chatting. Guiding by the complete video subtitles, the LLM is prompted to ask a series of more detailed questions, to uncover the underlying intentions. (2) To address the language incompleteness in both video segments and chatting dialogue, we ask an LLM to rewrite fluent and comprehensive descriptions that convey detailed user intent and what is occurring in the current video segment. (3) To handle the content fragmentation in video segments, we fine-tuned a PLM by prompting it to assess the relevance between two segments in a video, thereby supporting the searching of context and obtaining integrated content. Based on existing datasets, we reconstruct three In-VAL datasets to simulate user interactions across multi-domain and multilingual scenarios. Experiments indicate that our proposed Ask2Loc framework outperforms traditional end-to-end and two-stage methods across all datasets. 

To sum up, our contributions are threefold:

(1) We propose a new task, named In-VAL, which involves simulating the process of multiple interactions between humans and videos using an AI system to achieve instructional visual answer localization. We reconstruct three In-VAL datasets across multi-domain and multilingual scenarios, providing a paradigm and data support for future research.

(2) We introduce a novel framework, named Ask2Loc, for solving the In-VAL task. Apart from existing end-to-end and two-stage methods, it leverages language models to implement three interactive modules: chatting, rewriting, and searching, to address issues including user intent ambiguity, video segment incompleteness, and video content fragmentation, respectively.

(3) Extensive experiments on the In-VAL datasets demonstrate that our proposed Ask2Loc achieves an improvement by up to 14.91 in mIoU on the In-VAL task compared to the traditional end-to-end and two-stage methods.

\section{Related Work}
\subsection{Visual Answer Localization}
Visual answer localization involves identifying the most accurate span within a video that answers a natural language question ~\cite{gupta2023dataset}. Current approaches can be categorized into end-to-end model training and two-stage frameworks. End-to-end methods focus on strategies for integrating visual and textual features, as well as the utilization of multimodal information from videos, such as frame images and their corresponding subtitles ~\cite{kusa2022dossier,weng2023visual,wang2024grounded,li2024towards}. In contrast, two-stage frameworks employ LLMs or VLMs to preprocess the input questions and videos, thereby extracting richer information. This includes retrieving subtitle context ~\cite{tan2025rag}, expanding the question ~\cite{zheng2024training}, and generating visual descriptions ~\cite{qu2024chatvtg}, allowing for more comprehensive features to be utilized in the localization phase. However, these methods fail to truly simulate natural interactions between humans and video content during the localization and thus do not effectively resolve the semantic issues.

\subsection{Interaction by Asking Questions}
Asking questions is a natural way of interaction that can be used to clarify initial inquiries and acquire further knowledge ~\cite{li2016learning, abdelghani2024gpt,sasson2025art}. With recent advancements in LLMs and VLMs, posing questions to models can address tasks across various scenarios. For instance, in dialogue systems, engaging in multiple rounds of questioning with users can uncover their true and deeper intentions, thereby providing answers that better align with user expectations ~\cite{zhao2025medrag, li2024mediq,10.1145/3503161.3548361}. Additionally, in certain QA and content generation systems, using question-based prompts to rewrite user requests can expand input information and alleviate semantic discrepancies between input and retrieved content ~\cite{liu2024query,shu2024rewritelm,ma2023query}. Inspired by these studies, our In-VAL task involves interactively locating instructional visual answers, which requires addressing issues including intent ambiguity, description incompleteness, and content fragmentation. Leveraging LLMs and PLMs by asking questions can facilitate the implementation of these interactive processes to obtain better visual answers.

\section{Proposed Ask2Loc Framework}
In this section, we introduce Ask2Loc, our proposed interactive visual answer localization framework via asking questions. Formally, given an initial text question $Q$ provided by a user, and the intermediate further knowledge $K$ acquired by multiple interactions between the In-VAL system and a video $V$, our goal is to extract a video frame span from $T_s$ to $T_e$ so that the corresponding video clip can accurately answer the initial question $Q$.

\begin{figure}[htbp]
  \centering
  \includegraphics[width=.48\textwidth]{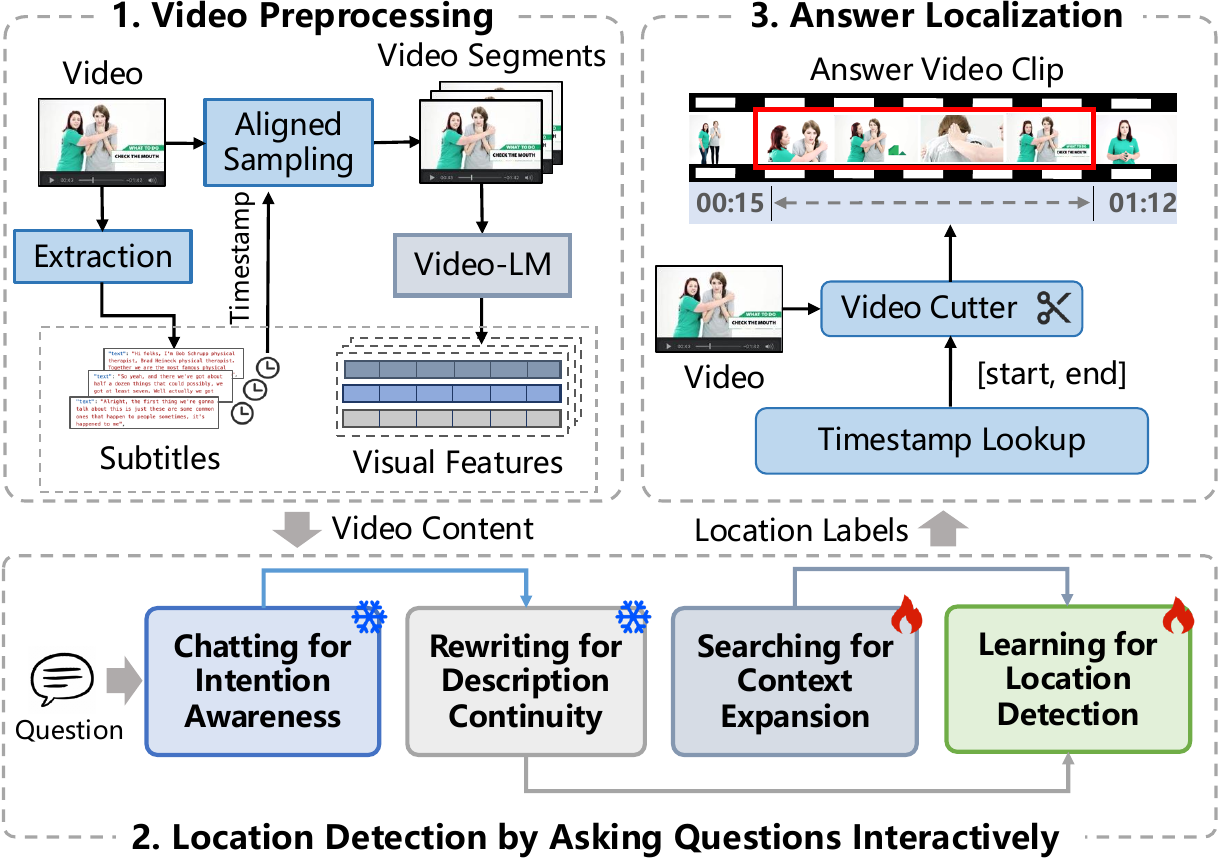}
  \caption{Ask2Loc utilizes the extracted subtitles and visual features from a video and learns to detect locations by leveraging language models and asking questions through three interactive modules. The predicted locations further facilitate cutting the video to obtain the final video clip.}
  \label{framework}
\end{figure}

The overall process of Ask2Loc is illustrated in Figure \ref{framework}, consisting of three primary phases: (1) video preprocessing to extract subtitles and the corresponding visual features from the aligned video segments, (2) location detection by asking questions through four interactive modules (the core phase), (3) answer localization by cutting the video with the predicted locations. We now describe each of these phases and modules in detail.

\begin{figure*}[htbp]
  \centering
  \includegraphics[width=.95\textwidth]{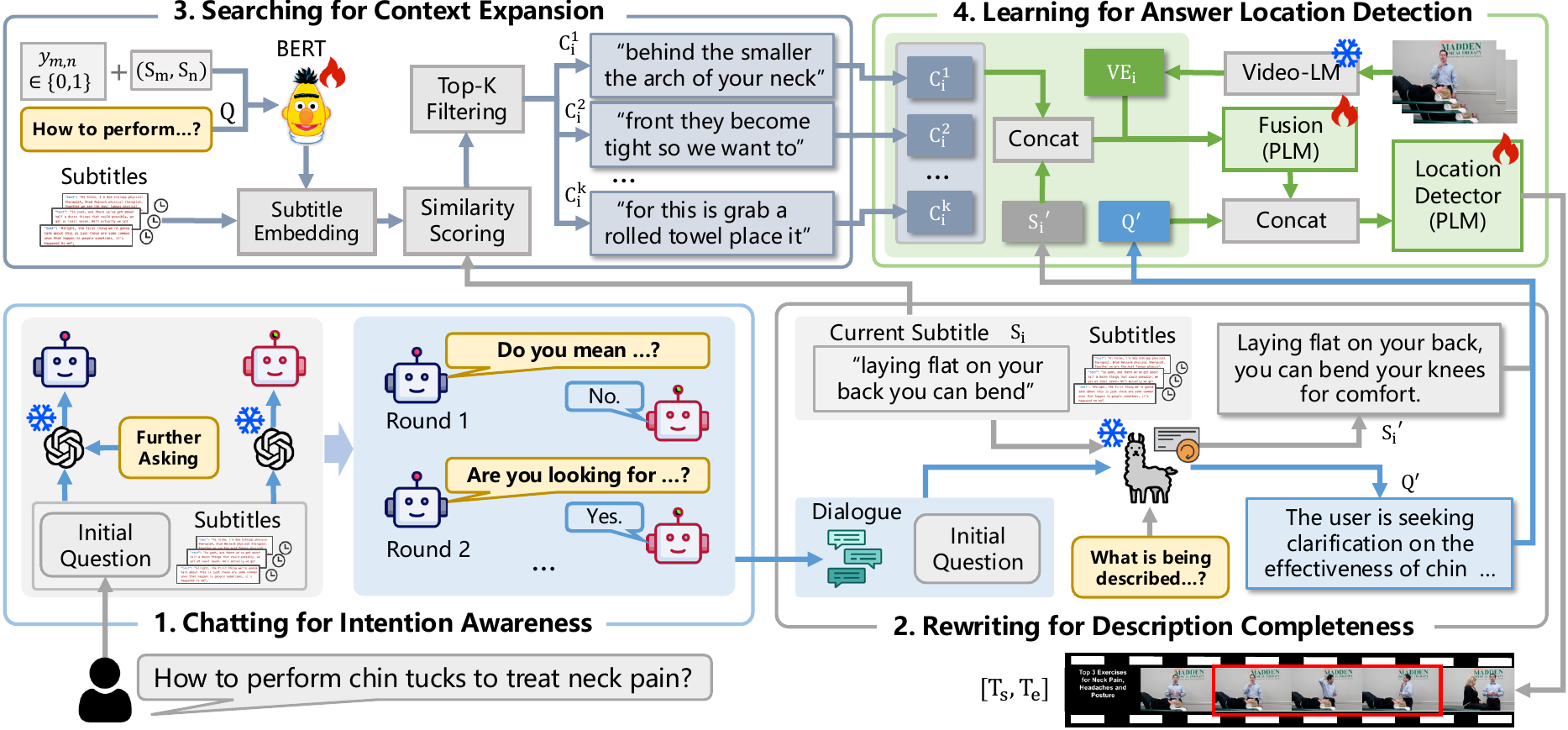}
  \caption{The detailed interactive and learning process of our proposed Ask2Loc. Three types of interactions are performed between language models and video content, including chatting for intention awareness, rewriting for description completeness, and searching for context expansion. The generated further knowledge, along with the corresponding visual features, are fused to detect the location of each video segment.}
  \label{method}
\end{figure*}

\subsection{Video Preprocessing}
Before performing In-VAL, we need to obtain the multimodal content of the video, which primarily includes the textual subtitles and the corresponding visual features. We first extract subtitles from the raw video files of existing datasets (MedVidQA ~\cite{gupta2023dataset}, VehicleVQA ~\cite{luo2019integrating}, and CMIVQA ~\cite{li2023overview}), and then perform merging and deduplication to obtain high-quality and complete subtitles. We then conduct video segment sampling \cite{zhang2023multi} from the original videos to align with the starting and ending timestamps of each subtitle $S_i$. For each video segment, we employ LLaVA-NeXT-Video ~\cite{lillava}, a modern visual-language model, to extract the embedding from the last hidden layer by prompting it with "\textit{What is shown in this video?}" to obtain the aligned visual feature for each video segment, denoted as $VE_i$.

\subsection{Location Detection by Asking Questions}

Inspired by the nature of human interactions with videos, and asking questions for clarification and knowledge acquisition from previous studies, we implement three modules to solve the aforementioned issues while performing interactive visual answer localization. Specifically, the three modules include: (1) chatting for intention awareness, (2) rewriting for description completeness, and (3) searching for context expansion. The three modules are particularly designed to address issues of intent ambiguity, language incompleteness, and content fragmentation, respectively. After conducting these interactive subtasks, a learning module is implemented for multimodal fusion and location detection. This results in the location of each video segment (within or outside the answer). The detailed process of this module is shown in Figure \ref{method}. We will provide a detailed explanation for each module.

\subsubsection{Chatting for Intension Awareness.} 
Instructional videos usually contain a large amount of domain knowledge that users have not yet mastered. Consequently, the initial questions posed by users are often vague in intention. This requires the system to be aware of the user intent through interactions and to formulate refined questions. Additionally, chatting is an efficient way of information exchange. In dialogue systems based on LLMs, asking users a series of follow-up questions during the conversation can help discern true intentions ~\cite{zhao2025medrag, li2024mediq,10.1145/3503161.3548361}, and finally aids in providing more accurate answers. Based on these observations, we attempt to simulate chatting interactions to gradually approach user intentions by employing LLMs.

Specifically, we prompt an LLM (GPT-4o\footnote{https://openai.com/index/gpt-4o-system-card/}) as the questioning agent by asking it to generate a further detailed question $Q_r$ in the chatting round $r$, referring to the initial question $Q$, the complete subtitles $S_1, S_2 ,..., S_n$, and all historical dialogue $D_{r-1}$ from previous $r-1$ rounds. Meanwhile, we establish another agent with the same LLM to simulate a user during application, by providing responses corresponding to each further question. To facilitate human users in potential future applications, we design a Boolean QA format. The agent poses questions that require only a "\textit{yes}" or "\textit{no}" answer, and the other agent provides one of these two options as the response. Through this chatting procedure, the semantics of an initial question are progressively aligned with the targeted video, resulting in a deepened and clarified user input. The above process can be formalized as:

\begin{small}
\begin{gather}
 Q_r = LLM(Q, S_1, S_2,...,S_n, D_{r-1},\texttt{Prompt}_q), \\
  A_r = LLM(Q_r, S_1, S_2,...,S_n, \texttt{Prompt}_a),
\label{eq:eq1}
\end{gather}
\end{small} where $Q_r$ and $A_r$ are QA pairs generated in the $r$-th round with the historical dialogue $D_{r-1}$, $\texttt{Prompt}_q$ and $\texttt{Prompt}_a$ are prompts for asking further questions and answering, respectively.

\subsubsection{Rewriting for Description Completeness.}
There are two aspects of semantic incompleteness in the process of In-VAL. First, for each video segment, the incomplete subtitle language leads to difficulties in understanding what is happening currently. Second, there remains a semantic gap between the QA dialogue generated from the previous chatting module and the actual intent description. We solve the above problems by employing rewriting with an LLM. LLM-based rewriting is an effective method to alleviate semantic discrepancies between users and systems ~\cite{liu2024query,shu2024rewritelm,ma2023query}. By asking an LLM with rewriting prompts, we implement a subtask to achieve language completeness in both user input and video content.

Specifically, for describing user input, we prompt an LLM (Llama 3.1-8B~\cite{grattafiori2024llama}) with "\textit{What the user want to ask according to the dialogue?}". By providing both the historical dialogue $D_r$ and the initial question $Q$, the LLM can summarize the complete information and generate a fluent description $Q'$ that contains the clarified user intent. Meanwhile, for each video segment, instead of using the incomplete subtitle directly, we perform rewriting by prompting an LLM with "\textit{What is being described in the current video segment according to the complete video subtitles?}". By providing the complete subtitles of the video as a reference, a fluent and concise description $S'$ is generated to replace the original subtitle $S$. Due to the need to call numerous APIs to rewrite a large number of subtitles, we opt
for the offline deployment of an LLM to accomplish this subtask. The above process can be formulated as:

\begin{small}
\begin{gather}
 Q' = LLM(Q, D_r,\texttt{Prompt}_{re\_q}), \\
  S_i' = LLM(S_i, S_1, S_2,...,S_n, \texttt{Prompt}_{re\_s}),
\label{eq:eq2}
\end{gather}
\end{small} where $Q'$ is the rewritten question and $S_i'$ is the descriptoin according to the $i$-th subtitle $S_i$, $\texttt{Prompt}_{re\_q}$ and $\texttt{Prompt}_{re\_s}$ are prompts for question rewriting and subtitle rewriting, respectively.

\subsubsection{Searching for Context Expansion}
Humans identify relevant video segments to aid in the assessment of the current one while performing localization. To simulate the process of relevant segment searching, we assume that for a given question, video segments within the corresponding answer span exhibit stronger correlations. Based on this hypothesis, we aim to leverage a pre-trained language model (PLM) to identify any other segments that are similar to a particular video segment, thereby expanding the context of the current content to further enhance understanding and assessment. Inspired by the embedding search in retrieval-augmented generation (RAG) systems ~\cite{asai2023self,jiang2023active,lewis2020retrieval}, we achieve a relevancy searching process for context expansion by tuning a PLM.

Specifically, we employ a PLM (BERT-Base ~\cite{devlin2019bert}) to calculate the similarity score between the current subtitle $S_i$ and each other subtitle $S_j$ from the same video. By concatenating the input question $Q_i$ and the targeted subtitle, denoted as $[Q; S_i]$, the question can guide the determination of relevance. To better adapt the model to the instructional domain, we construct a set of labeled subtitle pairs through random sampling for training and testing. By fine-tuning the PLM, we enable the model to better distinguish differences of subtitles within (labeled 1) and outside (labeled 0) the answer span. The cross-entropy tuning objective is formulated as below:

\begin{scriptsize}
\begin{gather}
\mathcal{L}_{s} = -\frac{1}{N}\sum_{m,n=1}^N [y_{m,n} log(\hat{y}_{m,n}) + (1 - y_{m,n})log(1 - \hat{y}_{m,n})], \\ 
\hat{y}_{m,n} = \sigma(W_s \cdot PLM([Q;S_m],S_n) + b_s),
\label{eq:eq3}
\end{gather}
\end{scriptsize} where $\hat{y}_{m,n} \in \{0,1\}$ is the predicted label of the text pair $[Q;S_m]$ and $S_n$ using $PLM$ and linear parameters $W_s$ and $b_s$. Then, each fragmented subtitle is embedded using the fine-tuned PLM. For a given subtitle, we perform a cosine similarity search using the embeddings, obtaining the top $k$ most similar segments as the context. This similarity scoring and filtering process can be formulated as:

\begin{footnotesize}
\begin{gather}
C_i^1,C_i^2,...C_i^k = Filter(Rank(Score_{i,1}, ... ,Score_{i,n}), k), \\
Score_{i,j} = Cosine(PLM(S_i,Q),PLM(S_j)),
\label{eq:eq4}
\end{gather}
\end{footnotesize} where $C_i^1,C_i^2,...C_i^k$ are filtered top-$k$ context by similarity ranking.

\subsubsection{Learning for Location Detection}
After acquiring the intent description and the content of video segments, we aim to use this further knowledge to detect whether each segment is within or outside the answer span. We define the detection as a classification task. Specifically, for each segment $i$, the visual feature ${VE}_i$ is firstly projected into the same space with textual features using linear parameters $W_v$ and $b_v$. The current description and its context are encoded using $PLM_{enc}$, the encoding layer of $PLM$. These two features are concatenated and fused with $PLM$. Then, a location classification is performed by integrating the fused feature $Fus_i$ with the feature from the intent description $Q'$. Given the true location and the predicted location of each segment as the label $L_i$ and $\hat{L}_i$, we fine-tune another $PLM$ for both modality fusion and location detection. The detection process can be formalized as:

\begin{scriptsize}
\begin{gather}
Fus_i = PLM\left([\sigma(W_v \cdot VE_i + b_v); PLM_{enc}([S'_i, C_i^1, ..., C_i^k])]\right) \\
\hat{L}_i = PLM([PLM_{enc}(Q'); Fus_i]).
\label{eq:eq5}
\end{gather}
\end{scriptsize} Our training objective is to minimize the cross-entropy loss between the predicted locations and the true labels for all of $N$ segments of a video. The overall loss can be formulated as:

\begin{small}
\begin{gather}
\mathcal{L} = - \frac{1}{N}\sum_{i=1}^N [L_i log \hat{L}_i + (1-L_i)log(1-\hat{L}_i)].
\label{eq:eq6}
\end{gather}
\end{small}

\subsection{Answer Localization}
Using the location detected for each video segment, we extract the answer video clip for the input question. During inference, for an instructional video, we gather all of its segments sequentially to perform the location detection as a batch. As a result, the segments of a video are labeled either within the answer (labeled 1) or outside the answer (labeled 0). We assume that the start position of the answer is the minimum timestamp of the segments labeled 1, and the end position is the maximum timestamp oppositely. Thus, we implement a timestamp lookup module to convert the predicted locations into a time span. This lookup process is denoted as follows.

\begin{small}
\begin{gather}
T_{start} = min(TS_{i} | L_i=1) \\ 
T_{end} = max(TS_{i} | L_i=1) + TD_{i},
\label{eq:eq7}
\end{gather}
\end{small} where $TS_{i}$ is the start timestamp of the $i$-th segment, and $TD_{i}$ is its duration time, $L_i=1$ is for retraining the segments within the answer span. Then, the answer video clip can be easily extracted from a video file through an open-source video-cutting tool.
\section{Experiments}
\subsection{Datasets and Data Construction}
To evaluate the performance of different methods in the In-VAL task, we reconstructed three datasets based on existing datasets to simulate user interactions during In-VAL. \textbf{MedVidQA} ~\cite{gupta2023dataset} is the dataset for answer localization in the medical domain that supports high-quality visual answers for textual questions. \textbf{VehicleVQA} ~\cite{luo2019integrating} is specifically designed for vehicle operation guidance and is created through processes that include video extraction, segmentation, and question annotation. \textbf{CMIVQA} ~\cite{li2023overview} is another dataset of medical instructional videos but with content in Chinese, which is then developed into a multilingual dataset ~\cite{li2024overview}. To construct In-VAL datasets (denoted as "In-X"), we assume that the ground-truth of each sample meets the expectations of users, and apply our interactive modules to the above datasets (more details in Appendix \ref{AppendixA}). We keep the multi-turn dialogue generated from the chatting module for each question. The rewritten user intent and current description for each video segment are also recorded, along with the top-k relevant subtitles obtained from the searching module. In addition, for each subtitle, we extract its aligned visual feature with a pre-trained VLM and save it as a tensor. Finally, a binary label of a segment is provided to represent whether the current position is within the answer span (see Appendix \ref{AppendixC} for a case). Table \ref{datacompare} presents the differences between our restructured In-VAL datasets and the traditional VAL datasets for each video segment. Table \ref{datastat} shows detailed statistics of these reconstructed datasets.
\begin{table*}[htbp]
\renewcommand{\arraystretch}{1.0}
\centering
\setlength{\tabcolsep}{2.5mm}
\caption{The differences between our reconstructed In-VAL datasets (highlighted in \sethlcolor{gray}\hl{gray}.) and existing VAL datasets. We populate additional information produced from multiple interactions to assist in localization in our In-VAL datasets.}
\resizebox{\linewidth}{!}{
\begin{tabular}{l|ccccccccc}\toprule
\multirow{1}{*}{\textbf{Dataset}} & \multicolumn{1}{c}{\makecell[c]{\textbf{Initial} \\ \textbf{Question}}} & \multicolumn{1}{c}{\makecell[c]{\textbf{Chatting} \\ \textbf{Dialogue}}} & \multicolumn{1}{c}{\makecell[c]{\textbf{Qestion} \\ \textbf{Description}}} & \multicolumn{1}{c}{\makecell[c]{\textbf{Video} \\ \textbf{Subtitle}}} & \multicolumn{1}{c}{\makecell[c]{\textbf{Subtitle} \\ \textbf{Description}}} & \multicolumn{1}{c}{\makecell[c]{\textbf{Subtitle} \\ \textbf{Context}}} & \multicolumn{1}{c}{\makecell[c]{\textbf{Visual} \\ \textbf{Feature}}} & \multicolumn{1}{c}{\makecell[c]{\textbf{Segment} \\ \textbf{Location}}} & \multicolumn{1}{c}{\makecell[c]{\textbf{Answer} \\ \textbf{Span}}}
\\ \midrule
\multirow{1}{*}{MedVidQA ~\cite{gupta2023dataset}} & \multicolumn{1}{c}{\ding{51}} & \multicolumn{1}{c}{\ding{55}} & \multicolumn{1}{c}{\ding{55}} & \multicolumn{1}{c}{\ding{51}} & \multicolumn{1}{c}{\ding{55}} & \multicolumn{1}{c}{\ding{55}} & \multicolumn{1}{c}{\ding{55}} & \multicolumn{1}{c}{\ding{55}} & \multicolumn{1}{c}{\ding{51}}
\\ 
\multirow{1}{*}{VehicleVQA ~\cite{luo2019integrating}} & \multicolumn{1}{c}{\ding{51}} & \multicolumn{1}{c}{\ding{55}} & \multicolumn{1}{c}{\ding{55}} & \multicolumn{1}{c}{\ding{51}} & \multicolumn{1}{c}{\ding{55}} & \multicolumn{1}{c}{\ding{55}} & \multicolumn{1}{c}{\ding{55}} & \multicolumn{1}{c}{\ding{55}} & \multicolumn{1}{c}{\ding{51}}
\\     
\multirow{1}{*}{CMIVQA ~\cite{li2023overview}} & \multicolumn{1}{c}{\ding{51}} & \multicolumn{1}{c}{\ding{55}} & \multicolumn{1}{c}{\ding{55}} & \multicolumn{1}{c}{\ding{51}} & \multicolumn{1}{c}{\ding{55}} & \multicolumn{1}{c}{\ding{55}} & \multicolumn{1}{c}{\ding{51}} & \multicolumn{1}{c}{\ding{55}} & \multicolumn{1}{c}{\ding{51}}
\\
\rowcolor{gray}
\multirow{1}{*}{\textbf{In-VAL Datasets (Ours)}} & \multicolumn{1}{c}{\ding{51}} & \multicolumn{1}{c}{\ding{51}} & \multicolumn{1}{c}{\ding{51}} & \multicolumn{1}{c}{\ding{51}} & \multicolumn{1}{c}{\ding{51}} & \multicolumn{1}{c}{\ding{51}} & \multicolumn{1}{c}{\ding{51}} & \multicolumn{1}{c}{\ding{51}} & \multicolumn{1}{c}{\ding{51}}
\\ \bottomrule
\end{tabular}}
\label{datacompare}
\end{table*}

\begin{table}[htbp]
\scriptsize
\renewcommand{\arraystretch}{1.1}
\centering
\setlength{\tabcolsep}{0.8mm}
\caption{Statistics of three In-VAL datasets. The number of videos (\#V), the number of questions (\#Q), the average video length ($\overline{\textbf{VL}}$), the average number of subtitle words ($\overline{\textbf{SW}}$), the average answer length ($\overline{\textbf{AL}}$), the number of training set (Train), and the number of test set (Test) are recorded.}
\resizebox{\linewidth}{!}{
\begin{tabular}{l|ccccccc}\toprule
\multirow{1}{*}{\textbf{Dataset}} & \multicolumn{1}{c}{\textbf{\#V}} & \multicolumn{1}{c}{\textbf{\#Q}} & \multicolumn{1}{c}{$\overline{\textbf{VL}}$} & \multicolumn{1}{c}{$\overline{\textbf{SW}}$} & \multicolumn{1}{c}{$\overline{\textbf{AL}}$} & \multicolumn{1}{c}{\textbf{Train}} & \multicolumn{1}{c}{\textbf{Test}}
\\ \midrule
\multirow{1}{*}{In-MedVidQA} & \multicolumn{1}{c}{846} & \multicolumn{1}{c}{2800} & \multicolumn{1}{c}{381.5} & \multicolumn{1}{c}{752.9} & \multicolumn{1}{c}{61.5} & \multicolumn{1}{c}{2660} & \multicolumn{1}{c}{140}
\\     
\multirow{1}{*}{In-VehicleQA} & \multicolumn{1}{c}{83} & \multicolumn{1}{c}{6743} & \multicolumn{1}{c}{90.2} & \multicolumn{1}{c}{476.8} & \multicolumn{1}{c}{25.0} & \multicolumn{1}{c}{5993} & \multicolumn{1}{c}{750}
\\     
\multirow{1}{*}{In-CMIVQA} & \multicolumn{1}{c}{1637} & \multicolumn{1}{c}{2937} & \multicolumn{1}{c}{257.4} & \multicolumn{1}{c}{535.4} & \multicolumn{1}{c}{34.3} & \multicolumn{1}{c}{2704} & \multicolumn{1}{c}{233}
\\ \bottomrule
\end{tabular}}
\label{datastat}
\end{table}

\subsection{Implementation Details}
For the chatting module, we use GPT-4o as the LLM to simulate a three-round dialogue to solve intent ambiguity. For the rewriting module, utilize Llama3.1-8B ~\cite{grattafiori2024llama} and Llama3-8B-Chinese ~\cite{shenzhi_wang_2024} to generate complete descriptions of the input question and the current video segment (see Appendix \ref{AppendixB} for specific prompts). For the searching module, we fine-tune the BERT ~\cite{devlin2019bert} and Erlangshen-RoBERTa ~\cite{fengshenbang} as embedding models for 2 epochs to learn the relevancy between any two subtitles and select the top 3 most relevant subtitles as the context of the current segment. We regard the localization learning process as a binary classification task. In the preprocessing step, each video segment is encoded using LLaVA-Next-Video ~\cite{lillava}. These visual embeddings and the corresponding textual features are fused using ModernBERT ~\cite{warner2024smarter}. The integrated features are further concatenated with the feature from the rewritten question to learn location detection using another ModernBERT model. We train the model for 8 epochs with the learning rate of 5e-5 and select the best result as the final performance.

\subsection{Evaluation Metrics}
Following previous studies ~\cite{gupta2023dataset,zhang2020span,yuan2019find}, we employ "R@n, IoU = $\mu$" and "mIoU" as evaluation metrics, treating the task of visual answer localization as a span prediction task. "R@n, IoU = $\mu$" measures the Intersection over Union (IoU) of the predicted visual span against the ground truth, considering only top-n retrieved moments where the overlap exceeds $\mu$. "mIoU" represents the mean IoU across all $N$ test samples. In our experiments, we set n = 1 and use $\mu$ values of 0.3, 0.5, and 0.7. The "mIoU" metric can be calculated as follows:

\begin{small}
\begin{equation}
\begin{aligned}
mIoU = \frac{1}{N} \sum_{i=1}^N \frac{A_i \cap B_i}{A_i  \cup B_i},
\label{eq:eq8}
\end{aligned}
\end{equation}
\end{small} where $A_i$ and $B_i$ represent the predicted span and the groundtruth span of the $i$-th sample, respectively.
\subsection{Compared Methods}
We evaluate our proposed Ask2Loc by comparing it with two frameworks. The first is named \textbf{End-to-End}, which directly provides answer spans based on the input question and video through an end-to-end inference or learning process. This framework includes the following representative methods: (1) Randomly guessing the answer location ~\cite{gupta2023dataset}, denoted as \textbf{RandomGuess}. (2) Using an LLM ~\cite{grattafiori2024llama} to directly generate answer locations from video subtitles, denoted as \textbf{LLM-Gen}. (3) Prompting a VLM ~\cite{zhang2024llavanextvideo} to generate answer locations from sampled frames of a video, denoted as \textbf{Video-LLM} (4) Predicting the answer position by integrating visual and textual features with a PLM (ModernBERT) ~\cite{zhang2021natural,wang2024grounded}, denoted as \textbf{PLM-Fusion}. (5) Predicting the location using subtitles enriched by their neighboring context (window size of 5) with the same PLM ~\cite{kusa2022dossier}, denoted as \textbf{PLM-Context}. (6) Using the integration of visual and textual features as soft prompts for answer spans prediction via the same PLM ~\cite{li2024towards,weng2023visual}, denoted as \textbf{PLM-Prompt}. The second framework is called \textbf{Two-Stage}, which utilizes the capabilities of a (visual) language model to preprocess textual or visual information to support the following localization. This framework includes (1) retrieving text segments related to the question with an PLM ~\cite{devlin2019bert} before localizing the answer ~\cite{tan2025rag}, denoted as \textbf{Retrieval-Loc}, (2) Rephrasing sub-questions contained within the initial question using an LLM ~\cite{grattafiori2024llama} before localizing the answer ~\cite{zheng2024training}, denoted as \textbf{Rephrase-Loc}. (3) Describing visual information in language using a VLM ~\cite{lillava} to match the input question before localizing the answer ~\cite{qu2024chatvtg}, denoted as \textbf{Desc-Loc}. 

\subsection{Main Results}
Comparative results of performance in In-VAL are shown in Table \ref{performance}.
\begin{table*}[htbp]
\renewcommand{\arraystretch}{1.1}
\centering
\setlength{\tabcolsep}{0.8mm}
\caption{Performance comparison of instructional visual answer localization for three types of frameworks and their representative methods. The best performance is highlighted in \sethlcolor{blue}\hl{blue}, while the second-best results are in \sethlcolor{green}\hl{green}. Our proposed Ask2Loc demonstrates performance improvements (highlighted in bold) across all three datasets and metrics.}
\resizebox{\linewidth}{!}{
\begin{tabular}{l|ccccc|cccc|cccc}\toprule
\multirow{2}{*}{\textbf{Framework}} & \multirow{2}{*}{\textbf{Method}} & \multicolumn{4}{|c}{\textbf{In-MedVidQA ~\cite{gupta2023dataset}}} & \multicolumn{4}{|c}{\textbf{In-VehicleVQA ~\cite{luo2019integrating}}} & \multicolumn{4}{|c}{\textbf{In-CMIVQA ~\cite{li2023overview}}}
\\ 
& \multicolumn{1}{l}{} & \multicolumn{1}{|c}{IoU=0.3} & \multicolumn{1}{c}{IoU=0.5} & \multicolumn{1}{c}{IoU=0.7} & \multicolumn{1}{c}{mIoU} & \multicolumn{1}{|c}{IoU=0.3} & \multicolumn{1}{c}{IoU=0.5} & \multicolumn{1}{c}{IoU=0.7} & \multicolumn{1}{c}{mIoU} & \multicolumn{1}{|c}{IoU=0.3} & \multicolumn{1}{c}{IoU=0.5} & \multicolumn{1}{c}{IoU=0.7} & \multicolumn{1}{c}{mIoU}
\\ \midrule
\multirow{6}{*}{End-to-End} & \multirow{1}{*}{RandomGuess ~\cite{gupta2023dataset}} & \multicolumn{1}{|c}{7.74} & \multicolumn{1}{c}{3.22} & \multicolumn{1}{c}{0.64} & \multicolumn{1}{c}{5.96} & \multicolumn{1}{|c}{6.12} & \multicolumn{1}{c}{1.89} & \multicolumn{1}{c}{1.28} & \multicolumn{1}{c}{4.96} & \multicolumn{1}{|c}{8.09} & \multicolumn{1}{c}{2.05} & \multicolumn{1}{c}{0.14} & \multicolumn{1}{c}{4.21}
\\
{} & \multirow{1}{*}{LLM-Gen ~\cite{grattafiori2024llama}} & \multicolumn{1}{|c}{15.07} & \multicolumn{1}{c}{6.88} & \multicolumn{1}{c}{2.79} & \multicolumn{1}{c}{9.42} & \multicolumn{1}{|c}{13.95} & \multicolumn{1}{c}{8.44} & \multicolumn{1}{c}{3.48} & \multicolumn{1}{c}{10.17} & \multicolumn{1}{|c}{13.27} & \multicolumn{1}{c}{5.69} & \multicolumn{1}{c}{0.80} & \multicolumn{1}{c}{8.57}
\\
{} & \multirow{1}{*}{Video-LLM ~\cite{zhang2024llavanextvideo}} & \multicolumn{1}{|c}{20.58} & \multicolumn{1}{c}{8.77} & \multicolumn{1}{c}{3.46} & \multicolumn{1}{c}{12.90} & \multicolumn{1}{|c}{19.01} & \multicolumn{1}{c}{11.85} & \multicolumn{1}{c}{4.42} & \multicolumn{1}{c}{14.65} & \multicolumn{1}{|c}{16.79} & \multicolumn{1}{c}{6.26} & \multicolumn{1}{c}{1.71} & \multicolumn{1}{c}{9.45}
\\
{} & \multirow{1}{*}{PLM-Fusion ~\cite{zhang2021natural,wang2024grounded}} & \multicolumn{1}{|c}{39.41} & \multicolumn{1}{c}{22.69} & \multicolumn{1}{c}{13.80} & \multicolumn{1}{c}{29.36} & \multicolumn{1}{|c}{49.58} & \multicolumn{1}{c}{\second{36.17}} & \multicolumn{1}{c}{19.77} & \multicolumn{1}{c}{38.26} & \multicolumn{1}{|c}{35.72} & \multicolumn{1}{c}{\second{20.46}} & \multicolumn{1}{c}{9.09} & \multicolumn{1}{c}{22.70}
\\
{} & \multirow{1}{*}{PLM-Context ~\cite{kusa2022dossier}} & \multicolumn{1}{|c}{48.22} & \multicolumn{1}{c}{28.57} & \multicolumn{1}{c}{14.85} & \multicolumn{1}{c}{35.46} & \multicolumn{1}{|c}{48.39} & \multicolumn{1}{c}{34.51} & \multicolumn{1}{c}{\second{21.10}} & \multicolumn{1}{c}{36.88} & \multicolumn{1}{|c}{33.91} & \multicolumn{1}{c}{17.04} & \multicolumn{1}{c}{10.51} & \multicolumn{1}{c}{20.26}
\\
{} & \multirow{1}{*}{PLM-Prompt ~\cite{li2024towards,weng2023visual}} & \multicolumn{1}{|c}{\second{52.14}} & \multicolumn{1}{c}{30.45} & \multicolumn{1}{c}{15.76} & \multicolumn{1}{c}{\second{37.08}} & \multicolumn{1}{|c}{\second{52.91}}& \multicolumn{1}{c}{34.81} & \multicolumn{1}{c}{20.74} & \multicolumn{1}{c}{\second{40.37}} & \multicolumn{1}{|c}{\second{39.54}} & \multicolumn{1}{c}{19.12} & \multicolumn{1}{c}{\second{14.17}} & \multicolumn{1}{c}{\second{27.52}}
\\ \midrule 
\multirow{3}{*}{Two-Stage}  & \multirow{1}{*}{Retrieval-Loc ~\cite{tan2025rag}} & \multicolumn{1}{|c}{35.25} & \multicolumn{1}{c}{21.47} & \multicolumn{1}{c}{10.81} & \multicolumn{1}{c}{28.52} & \multicolumn{1}{|c}{39.04} & \multicolumn{1}{c}{28.36} & \multicolumn{1}{c}{18.15} & \multicolumn{1}{c}{33.72} & \multicolumn{1}{|c}{22.86} & \multicolumn{1}{c}{12.41} & \multicolumn{1}{c}{4.77} & \multicolumn{1}{c}{16.59}
\\
{} & \multirow{1}{*}{Expand-Loc ~\cite{zheng2024training}} & \multicolumn{1}{|c}{36.83} & \multicolumn{1}{c}{24.50} & \multicolumn{1}{c}{7.25} & \multicolumn{1}{c}{29.47} & \multicolumn{1}{|c}{41.62} & \multicolumn{1}{c}{26.75} & \multicolumn{1}{c}{13.18} & \multicolumn{1}{c}{32.14} & \multicolumn{1}{|c}{19.56} & \multicolumn{1}{c}{12.37} & \multicolumn{1}{c}{2.68} & \multicolumn{1}{c}{15.81}
\\
{} & \multirow{1}{*}{Describe-Loc ~\cite{qu2024chatvtg}} & \multicolumn{1}{|c}{42.61} & \multicolumn{1}{c}{\second{30.65}} & \multicolumn{1}{c}{\second{19.76}} & \multicolumn{1}{c}{31.02} & \multicolumn{1}{|c}{33.38} & \multicolumn{1}{c}{24.36} & \multicolumn{1}{c}{16.01} & \multicolumn{1}{c}{26.75} & \multicolumn{1}{|c}{30.81} & \multicolumn{1}{c}{15.02} & \multicolumn{1}{c}{8.07} & \multicolumn{1}{c}{20.76}
\\ \midrule
{Interactive} & \multirow{1}{*}{\textbf{Ask2Loc (ours)}} & \multicolumn{1}{|c}{\best{62.78}} & \multicolumn{1}{c}{\best{33.37}} & \multicolumn{1}{c}{\best{23.57}} & \multicolumn{1}{c}{\best{43.22}} & \multicolumn{1}{|c}{\best{67.42}} & \multicolumn{1}{c}{\best{52.65}} & \multicolumn{1}{c}{\best{35.54}} & \multicolumn{1}{c}{\best{55.28}} & \multicolumn{1}{|c}{\best{42.46}} & \multicolumn{1}{c}{\best{25.04}} & \multicolumn{1}{c}{\best{17.14}} & \multicolumn{1}{c}{\best{31.37}}
\\ \midrule
\multicolumn{2}{c}{Improvement over 2nd-best } & \multicolumn{1}{|c}{\textbf{+10.64}} & \multicolumn{1}{c}{\textbf{+2.72}} & \multicolumn{1}{c}{\textbf{+3.81}} & \multicolumn{1}{c}{\textbf{+6.14}} & \multicolumn{1}{|c}{\textbf{+14.51}} & \multicolumn{1}{c}{\textbf{+16.48}} & \multicolumn{1}{c}{\textbf{+14.44}} & \multicolumn{1}{c}{\textbf{+14.91}} & \multicolumn{1}{|c}{\textbf{+2.92}} & \multicolumn{1}{c}{\textbf{+4.58}} & \multicolumn{1}{c}{\textbf{+2.97}} & \multicolumn{1}{c}{\textbf{+3.85}}
\\ \bottomrule
\end{tabular}}
\label{performance}
\end{table*}
Overall, our Ask2Loc demonstrates advantages across all three datasets. As our In-VAL dataset is constructed by simulating user interactions, these results demonstrate that Ask2Loc can better meet user expectations by utilizing additional information generated during multiple interactions. Specifically, compared to End-to-End and Two-Stage methods, Ask2Loc achieves more accurate visual answer localization across all metrics. In the mIoU metric, it achieves an improvement of 6.14 in the English medical domain (In-MedVidQA), an improvement of 14.91 in the automotive guidance domain (In-VehicleVQA), and an improvement of 3.85 in the Chinese medical domain (In-CMIVQA).

Meanwhile, compared to Two-Stage methods, End-to-End methods undergo integrated training for the entire process from input to output. They naturally avoid errors that may arise during the intermediate stages of data augmentation, thereby achieving better overall performance than Two-Stage methods. In contrast, our interactive framework, by incorporating more comprehensive information of videos and users, mitigates error dependency in intermediate stages and allows subsequent training phases to benefit from the previously generated knowledge. This demonstrates that our framework can effectively balance knowledge enhancement and model training, resulting in improved performance.

Furthermore, different frameworks and methods vary widely in performance across domains. Overall, the In-VAL task seems relatively easier on In-MedVidQA (English medical domain) and In-VehicleVQA (English vehicle domain) datasets, whereas it proves more challenging on In-CMIVQA (Chinese medical domain). Our method, compared to baselines, achieves a tremendous improvement on the In-CMIVQA dataset, demonstrating its capability to handle challenging scenarios. On the other hand, our method achieves the best results on the In-VehicleVQA dataset and performs greatly better than on other datasets, indicating the superiority of our method in handling different scenarios.

\subsection{In-Depth Study}
\subsubsection{Ablative Results}
We first study the contribution of each top-level module in our Ask2Loc. Table \ref{ablation-top} illustrates the impact on overall performance when one of the three interactive modules is removed from the framework (denoted as "w/o X"). We notice that each interactive module positively contributes to the overall performance of answer localization. On different datasets, the impact of each module varies. The Rewriting module has a more pronounced effect on MedVidQA and CMIVQA, whereas the Searching module is more beneficial for VehicleQA. The ablation of the Chatting module leads to substantial performance loss on all three datasets.
\begin{table}[htbp]
\renewcommand{\arraystretch}{1.1}
 \centering
 \setlength{\tabcolsep}{0.8mm}
  \caption{Ablative study on top-level interactive modules. mIoU results indicate that evey module enhances overall performance. The largest impacts are highlighted in \sethlcolor{blue}\hl{blue}.}
 \resizebox{\linewidth}{!}{
 \begin{tabular}{p{0.3\linewidth}|ccc}\toprule
    \textbf{Method} & \textbf{In-MedVidQA} & \textbf{In-VehicleVQA} & \textbf{In-CMIVQA} \\
    \midrule \specialrule{0em}{1.5pt}{1.5pt}
    Ask2Loc & 43.22 & 55.28 & 31.37 \\
    \midrule
    - w/o Chatting & 39.18 & 45.53 & 24.67 \\
    - w/o Rewriting & \best{37.08} & 50.64 & \best{21.88} \\
    - w/o Searching & 41.48 & \best{37.83} & 28.17 \\
    \bottomrule
 \end{tabular}}
 \label{ablation-top}
\end{table}

We then dive into the analysis of specific modules. Within the rewriting module, we evaluate impacts of rewriting subtitles and input questions on overall performance, denoted as "w/o S-Rewriting" and "w/o Q-Rewriting", respectively. For the searching module, we analyze the impact of fine-tuning PLM on performance by replacing the updated PLM with its original version, referred to as "w/o FT-Searching". For the detection module, we remove the fusion component to evaluate the role of visual features, denoted as "w/o Vis-Detector". The mIoU results shown in Table \ref{ablation-mod} indicate that all these detailed settings contribute to the overall performance.
\begin{table}[htbp]
\renewcommand{\arraystretch}{1.1}
 \centering
 \setlength{\tabcolsep}{0.8mm}
  \caption{Ablative study on detailed settings in specific modules of Ask2Loc. The largest impacts are highlighted in \sethlcolor{blue}\hl{blue}.}
 \resizebox{\linewidth}{!}{
 \begin{tabular}{p{0.35\linewidth}|ccc}\toprule
    \textbf{Method} & \textbf{In-MedVidQA} & \textbf{In-VehicleVQA} & \textbf{In-CMIVQA} \\
    \midrule \specialrule{0em}{1.5pt}{1.5pt}
    Ask2Loc & 43.22 & 55.28 & 31.37 \\
    \midrule
    - w/o S-Rewriting & 41.91 & 52.86 & \best{24.51} \\
    - w/o Q-Rewriting & \best{39.55} & 52.59 & 27.09 \\
    \midrule
    - w/o FT-Searching & 41.57 & \best{42.12} & 28.74 \\
    \midrule
    - w/o Vis-Detector & 40.10 & 45.92 & 27.93 \\
    \bottomrule
 \end{tabular}}
 \label{ablation-mod}
\end{table}

\subsubsection{Impact of Asking Prompt}
Prompt formatting has been proven to largely affect LLM performance in various tasks ~\cite{he2024does}. We further explore the impact of the asking prompts during interactions in our Ask2Loc. In contrast, we design a corresponding imperative prompt for each asking prompt. Specifically, for the chatting module, instead of making an LLM ask further questions, we prompt an LLM to generate further statements to directly explain the user intention given the video content and the answer span, denoted as "Chatting w/o Ask". For the rewriting module, instead of asking an LLM "\textit{What is happening now in the video?}" to conduct the subtitle rewriting, we provide the corresponding imperative prompt that conveys the same meaning: "\textit{Rewrite the subtitle to describe the content of the current clip}", denoted as "Rewriting w/o Ask". For the searching module, we directly remove the appended question for tuning the PLM to conduct a simple sentence pair classification, denoted as "Searching w/o Ask". As shown in Table \ref{prompt}, asking prompts can positively influence performance across all three modules compared to the corresponding imperative prompts. Notably, a larger effect is observed in the chatting and searching modules, where the largest improvements in mIoU are 4.21 and 3.52, respectively. This indicates that posing questions can encourage the model to better reflect on the current task and content, thereby facilitating improved responses, including uncovering user intent and learning content relevance between two video segments.

\begin{table}[htbp]
\renewcommand{\arraystretch}{1.1}
 \centering
 \setlength{\tabcolsep}{0.8mm}
  \caption{Study of prompt formatting in three interactive modules. The largest impacts are highlighted in \sethlcolor{blue}\hl{blue}.}
 \resizebox{\linewidth}{!}{
 \begin{tabular}{p{0.35\linewidth}|ccc}\toprule
    \textbf{Method} & \textbf{In-MedVidQA} & \textbf{In-VehicleVQA} & \textbf{In-CMIVQA} \\
    \midrule \specialrule{0em}{1.5pt}{1.5pt}
    Ask2Loc & 43.22 & 55.28 & 31.37 \\
    \midrule
    - Chatting w/o Ask & \best{39.01} & 52.12 & \best{28.75} \\
    - Rewriting w/o Ask & 42.74 & 53.95 & 30.21 \\
    - Searching w/o Ask & 40.47 & \best{51.76} & 29.59 \\
    \bottomrule
 \end{tabular}}
 \label{prompt}
\end{table}

\subsubsection{Impacts of Language Models}
We further investigate the impacts of different language models within each interactive module (see Appendix \ref{AppendixD} for impacts on learning localization). In the chatting module, we directly call LLM APIs for dialogue generation, using GPT-4o and GPT-4o-mini for comparison. In the rewriting module, due to the large volume of inference times, we utilize locally deployed models for comparison, including the English and Chinese versions of Llama3-8B and Qwen2.5-3B for both languages. In the searching module, we select different PLMs, including the English version and Chinese version of BERT-Base and RoBERTa-Base, denoted as "BERT (en/zh)" and "RoBERTa (en/zh)". Results in Table \ref{lm} indicate that performance and model capability exhibit a generally positive correlation.
\begin{table}[htbp]
\renewcommand{\arraystretch}{1.1}
 \centering
 \setlength{\tabcolsep}{0.8mm}
  \caption{Impacts of language models employed in each interactive module. Better performance is highlighted in \sethlcolor{blue}\hl{blue}.}
 \resizebox{\linewidth}{!}{
 \begin{tabular}{l|l|ccc}\toprule
    \textbf{Module} & \textbf{Language Model} & \textbf{In-MedVidQA} & \textbf{In-VehicleVQA} & \textbf{In-CMIVQA} \\
    \midrule \specialrule{0em}{1.5pt}{1.5pt}
    \multirow{2}{*}{Chatting} & GPT-4o & \best{43.22} & \best{55.28} & \best{31.37} \\
     & GPT-4o-mini & 42.46 & 53.92 & 29.58 \\
    \midrule
    \multirow{2}{*}{Rewriting} & Llama3-8B (en/zh) & \best{43.22} & \best{55.28} & \best{31.37} \\
     & Qwen2.5-3B & 41.60 & 53.17 & 30.79 \\
    \midrule
    \multirow{2}{*}{Searching} & RoBERTa (en/zh) & \best{43.33} & 54.75 & \best{31.37} \\
    & BERT (en/zh) & 43.22 & \best{55.28} & 28.95 \\
    \bottomrule
 \end{tabular}}
 \label{lm}
\end{table}

\subsubsection{Study of Hyperparameters}
\begin{figure}[htbp]
  \centering
  \includegraphics[width=.48\textwidth]{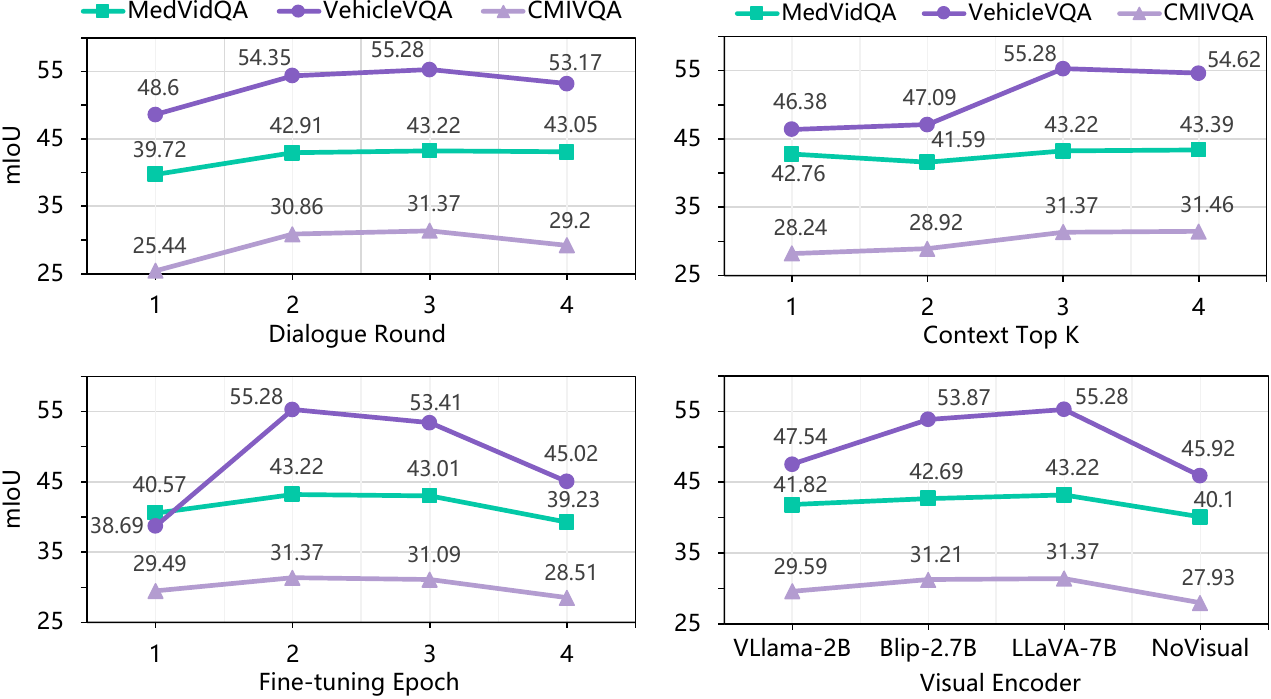}
  \caption{The impact of key hyperparameters, including the number of dialogue rounds in Chatting (top left), the number of top relevant subtitles in Searching (top right), the number of tuning epochs for relevance in Searching (bottom left), and the visual encoder models (bottom right).}
  \label{hyper}
\end{figure}

We evaluate the impact of important hyperparameters under different settings. For the Chatting module, we vary dialogue rounds ranging from 1 to 4. For the Searching module, we test the impact of top k relevant contexts on the results. We also explore fine-tuning epochs from 1 to 4 (with the same learning rate of 5e-5) for learning relevancy with a PLM. Finally, we evaluate different VLMs for extracting visual features, including LLaVA-Next-Video-7B ~~\cite{lillava} (LLaVA-7B), VideoLlama-2B ~\cite{zhang2025videollama} (VLlama-2B), and Blip2-2.7B ~\cite{li2023blip} (VBlip-2.7B), along with learning without visual features (NoVisual). As shown in Figure \ref{hyper}, we discover that simulating three rounds of dialogue is the most effective way to address user intent. Additionally, identifying the three most relevant subtitles can largely improve performance. Meanwhile, fine-tuning a PLM for two rounds allows better adaptation to domain-specific relevance assessments. Finally, a larger VLM is more conducive to integrate with textual features.
\subsection{Qualitative analysis}
\begin{figure}[htbp]
  \centering
  \includegraphics[width=.48\textwidth]{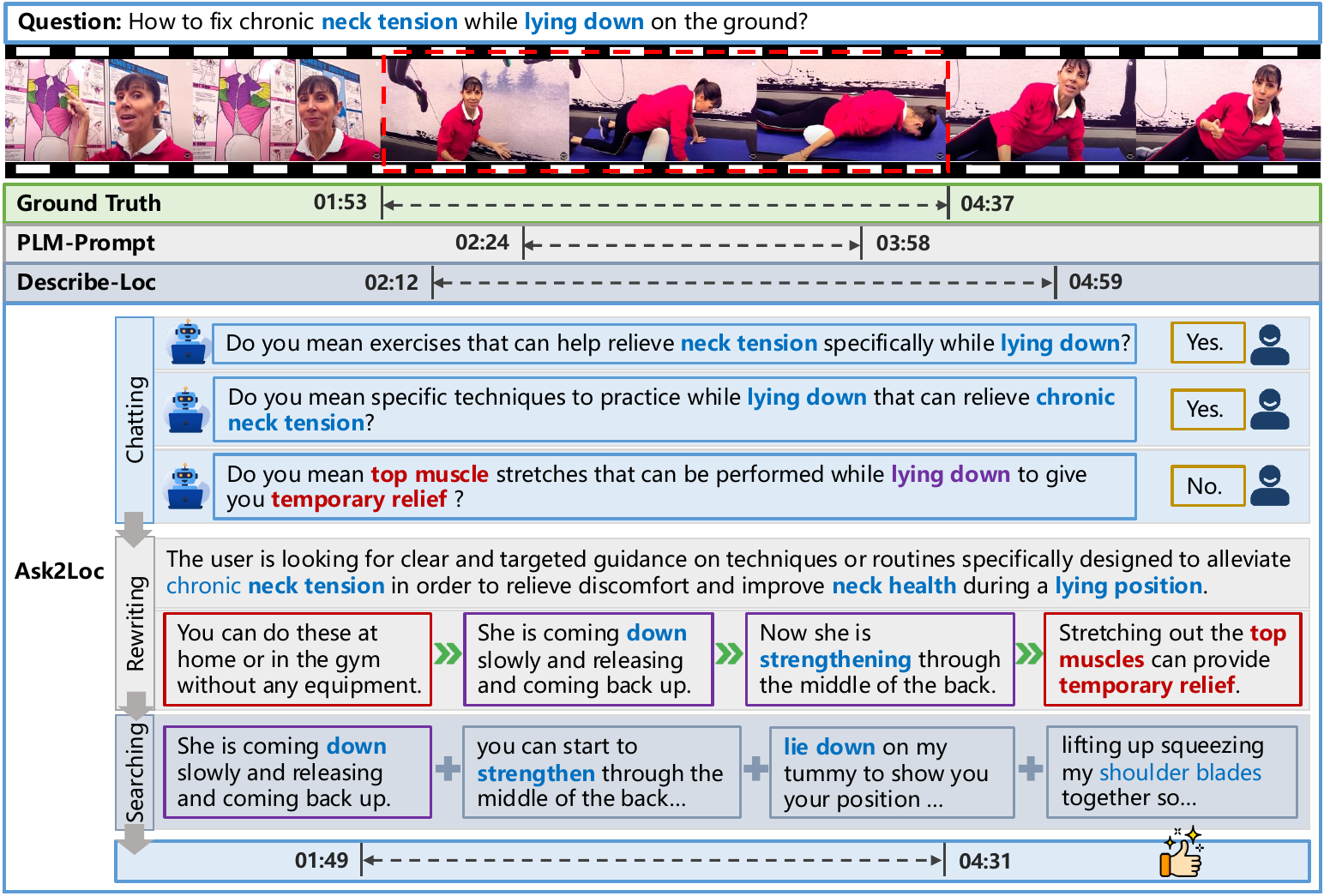}
  \caption{Qualitative analysis of our Ask2Loc framework on a case in In-MedVidQA. Compared to baselines, Ask2Loc simulates multiple interactions to facilitate a more expected and accurate answer location. Key supporting information is in \textcolor{darkblue}{blue text}, while misleading content is marked in \textcolor{red}{red text}.}
  \label{case}
\end{figure}

Figure \ref{case} illustrates an example from the In-MedVidQA dataset. Compared to two representative baseline methods (End-to-End PLM-Prompt and Two-Stage Describe-Loc), our Ask2Loc can accumulate richer information related to the input question and each video segment via three interactive modules, thereby supporting more effective visual answer localization. The compared methods, due to a lack of in-depth understanding and further expansion on individual segments, tend to provide answer spans that are either overly broad or excessively narrow. In comparison, Ask2Loc better simulates interactions performed by humans, thus presenting a localization capability more aligned with user expectations. More comparative cases are shown in Appendix \ref{AppendixE}. We also describe the interactions with users when applying Ask2Loc in Appendix \ref{AppendixF}.
\section{Conclusion}
We investigate the interactions between humans and videos during instructional visual answer localization and propose a new task, named In-VAL. To address issues including user intent ambiguity, video segment incompleteness, and video content fragmentation in this task, we introduce a novel framework, named Ask2Loc. It contains three asking-style interactive modules including chatting, rewriting, and searching to address these issues separately, thereby aiding in obtaining user-expected locations. We reconstruct three datasets for the In-VAL task across multi-domain and multilingual scenarios. Experiments on these datasets demonstrate that our Ask2Loc, by learning with comprehensive information introduced during multiple interactions, outperforms end-to-end and two-stage approaches. In the future, we will further explore more adaptive and generalized interactive processes in the In-VAL task.

{
    \small
    \bibliographystyle{ieeenat_fullname}
    \bibliography{main}
}

\clearpage
\maketitlesupplementary

\appendix
\section{User Expectation Simulation in In-VAL Task} \label{AppendixA}
In existing VAL tasks, systems directly derive the location of instructional visual answers based on the input questions. This approach overlooks the multiple interactions human users engage in with video content while searching for answer locations, potentially leading to semantically correct but user-unexpected answers. Under our In-VAL task paradigm, we emphasize the natural interactions between users and video content. We develop the In-VAL dataset by simulating human user interactions through LLMs and PLMs. Using the ground truth answers from the original VAL datasets as results that are both expected by users and semantically correct, the original data is served as the target to guide the reconstruction of our new interactive datasets. We introduce the generated data from the simulated interactions into our new In-VAL dataset, so that the dataset consists of additional knowledge including asking-formatted dialogues for user intent clarification, comprehensive descriptions of current video content, and contextual information of the current video segments, to further assist visual answer localization.

It should be noted that our In-VAL datasets does not include ground truth locations in the training and testing samples. Instead, the ground truth answers are used to guide the generation of dialogue content that better meets user expectations, simulating interactions between users and the system in real-world scenarios. The actual location labels are used separately for training and evaluation. Moreover, because the construction of our In-VAL datasets is guided by ground truth answers, the evaluation results on these datasets also reflect the ability to provide answers that align with user expectations.

\section{Prompts for Interactions} \label{AppendixB}
\subsection{Further Questioning for Chatting}
Ask2Loc utilizes an LLM to generate further questions, enabling a deeper understanding of user intent. The prompt for English scenarios is as follows:
\begin{systemprompt}
You are an AI assistant to help a user find out his/her true intention by asking questions.
\end{systemprompt}

\begin{userprompt}
Please ask the user a further question to better understand his/her intention, according to the following guidelines: 

1. You should generate a further question for the user referring to the his/her INITIAL\_QUESTION.

2. Your further question need to align with the content of the HISTORY\_DIALOGUE which contains question-answering pairs, each of which has a yes-or-no question followed by its answer.

3. Your further question should be more explict and better for screening out irrelevant video descriptions in the DESCRIPTION\_SPANS list.

4. Your further question should be a yes-or-no question for the user to only response "yes" or "no", you can start with "Do you mean ...".

INITIAL\_QUESTION: \{init\_question\}

HISTORY\_DIALOGUE: \{hist\_dialogue\}

DESCRIPTION\_SPANS: \{description\_spans\}

Your further question: 
\end{userprompt}

The corresponding prompt for Chinese scenarios is as follows:
\begin{systemprompt}
\begin{CJK}{UTF8}{gbsn}
你是一个AI助手能通过提出进一步的问题帮助用户找出他/她的真实意图。
\end{CJK}
\end{systemprompt}

\begin{userprompt}
\begin{CJK}{UTF8}{gbsn}
请参考以下指引提问用户一个进一步的问题来更好地理解他/她的意图： 

1. 你应该参考他/她的初始问题生成一个进一步的问题给用户。

2. 你提出的问题需要与历史对话内容一致，历史对话包含了问题与答案对，每一个问题有一个是或否的答案。

3. 你提出的问题应该有更加明确的意图并且能帮助更好地过滤掉描述列表中不相关的描述文本。

4. 你提出的问题应该是一个可以用“是”或“否”回答的问题，比如你可以用“你的意思是...？”来开头进行提问。   

初始问题：\{init\_question\}

历史对话： \{hist\_dialogue\}

描述列表： \{description\_spans\}

你的问题：
\end{CJK}
\end{userprompt}

\subsection{Yes-No Answering for Chatting}
For each generated further question during chatting, another LLM provides a corresponding "yes" or "no" answer to format a complete dialogue and clarify the user intent. The prompt for English scenarios is as follows.
\begin{systemprompt}
You are an AI assistant to answer a yes-or-no question referring to the information provided.
\end{systemprompt}

\begin{userprompt}
Please answer a question with "yes" or "no", according to the following guidelines: 

1. You should refer to the provided TEXT\_CONTENT to answer the QUESTION.

2. You should only output "yes" or "no" as the answer for the QUESTION.

QUESTION: \{question\}

TEXT\_CONTENT: \{text\}

Answer: 
\end{userprompt}

The corresponding prompt for Chinese scenarios is as follows:
\begin{systemprompt}
\begin{CJK}{UTF8}{gbsn}
你是一个AI助手能用是或否回答一个问题来帮助澄清用户的意图。
\end{CJK}
\end{systemprompt}

\begin{userprompt}
\begin{CJK}{UTF8}{gbsn}
请参考用户的初始问题回答一个针对它的进一步的问题。请以“是”或“否”回答，不要生成任何其他信息。

问题：\{question\}

参考的初始问题：\{initial\_question\}

你的回答：
\end{CJK}
\end{userprompt}

\subsection{Dialogue Summary for Rewriting}
Based on the user's initial question and the historical dialogue from the chatting module, an LLM is used to reinterpret the user's intent to provide comprehensive and in-depth request information. The prompt for English scenarios is as follows.
\begin{systemprompt}
You are an AI assistant to summarize a multi-turn dialogue that is related to a question from a user.
\end{systemprompt}

\begin{userprompt}
INITIAL QUESTION: \{question\}

DIALOGUE:
\{dialogue\}

According to the dialogue between an inquirer and a user about further clarify an initial question above, what the user really want to ask?
\end{userprompt}

The corresponding prompt for Chinese scenarios is as follows:
\begin{systemprompt}
\begin{CJK}{UTF8}{gbsn}
你是一个AI助手能总结跟一个问题相关的对话的内容。
\end{CJK}
\end{systemprompt}

\begin{userprompt}
\begin{CJK}{UTF8}{gbsn}
请总结提问者与用户之间关于澄清一个初始问题的对话。你应该输出一个关于用户问题的真正意图的描述。

初始问题：\{question\}

对话：

\{dialogue\}

你的描述：
\end{CJK}
\end{userprompt}

\subsection{Subtitle Description for Rewriting}
Similarly to the question rewriting, an LLM is prompted to describe what is happening in a specific video segment by referring to the complete subtitles of the video. The prompt for the English scenario is as follows.
\begin{systemprompt}
You are an AI assistant to describe what is currently happening given the complete subtitles of a video.
\end{systemprompt}

\begin{userprompt}
Complete Subtitles: \{all\_subtitles\}

Current Subtitle: \{subtitle\}

Referring to the complete subtitles from a video and the current subtitle above, what is happening now in the video?
\end{userprompt}

The corresponding prompt for Chinese scenarios is as follows:
\begin{systemprompt}
\begin{CJK}{UTF8}{gbsn}
你是一个AI助手能描述当前视频片段正在发生的内容。
\end{CJK}
\end{systemprompt}

\begin{userprompt}
\begin{CJK}{UTF8}{gbsn}
请参考以下视频完整字幕，对以下当前字幕进行重写，简要描述当前时刻正在发生什么。请直接用中文描述。

视频完整字幕：\{all\_subtitles\}

当前字幕：\{subtitle\}

你的描述：
\end{CJK}
\end{userprompt}

\section{Case of In-VAL Dataset} \label{AppendixC}
\begin{figure*}[htbp]
  \centering
  \includegraphics[width=.95\textwidth]{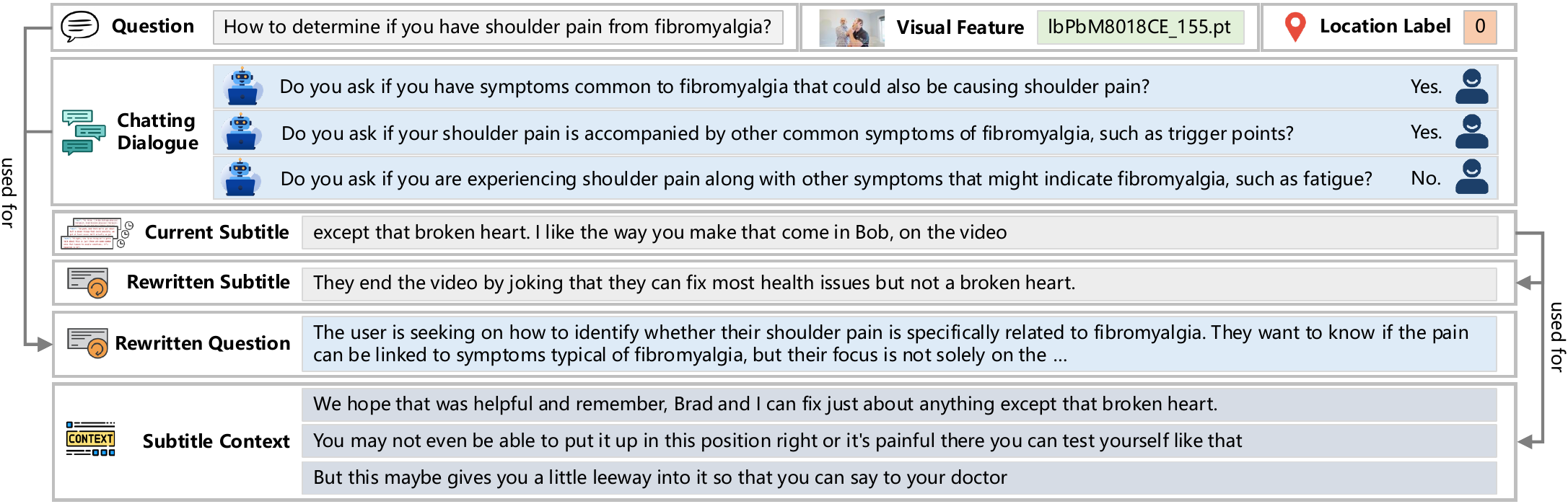}
  \caption{We present the reconstructed datasets for In-VAL. The multi-round (3 rounds in this case) dialogue is reserved for each input question. The rewritten question and subtitle for the current segment are also recorded, along with its relevant context from other segments (top 3 in this case).}
  \label{dataset}
  \vspace{-10pt}
\end{figure*}

We have reconstructed three In-VAL datasets. For each sample in the dataset (as shown in Figure \ref{dataset}), we simulate the process of multiple interactions between humans and video content, integrating the additional knowledge obtained from these interactions with the questions, subtitles, and other information from the original data. The In-VAL dataset offers a new paradigm for interactively locating visual answers in instructional videos.

\section{Impact of PLM on Location Learning} \label{AppendixD}
We further evaluate the impact of different PLMs on the performance of the location detection module. Table \ref{plm} presents mIoU results when using different PLMs for both English and Chinese datasets. Overall, the stronger the capability of the PLM, the better the final performance of Ask2Loc. However, even when using smaller or less capable PLMs, our framework still outperforms baseline methods.

\begin{table*}[htbp]
\renewcommand{\arraystretch}{1.1}
 \centering
 \setlength{\tabcolsep}{5mm}
  \caption{Impact of PLM for learning location detection in our Ask2Loc. Overall, the performance of location detection (mIoU) is positively correlated with the capabilities of the PLM.}
  \vspace{-5pt}
 \resizebox{\linewidth}{!}{
 \begin{tabular}{c|c|ccc}\toprule
    \textbf{Language} & \textbf{PLM} & \textbf{In-MedVidQA} & \textbf{In-VehicleVQA} & \textbf{In-CMIVQA} \\
    \midrule \specialrule{0em}{1.5pt}{1.5pt}
    EN & ModernBERT-Base ~\cite{warner2024smarter} (149M) & 43.22 & 55.28 & - \\
    EN & BERT-Base ~\cite{devlin2019bert} (110M) & 41.58 & 49.91 & - \\
    EN & DistilBERT-Base ~\cite{sanh2019distilbert} (66M) & 40.39 & 47.84 & - \\
    
    ZH & Erlangshen-RoBERTa ~\cite{fengshenbang} (110M) & - & - & 31.37 \\
    ZH & BERT-Base-Chinese ~\cite{devlin2019bert} (110M) & - & - & 30.09 \\
    \bottomrule
 \end{tabular}}
 \label{plm}
 \vspace{-5pt}
\end{table*}

\section{More Cases of Localization Performance} \label{AppendixE}
\begin{figure*}[htbp]
  \centering
  \includegraphics[width=.8\textwidth]{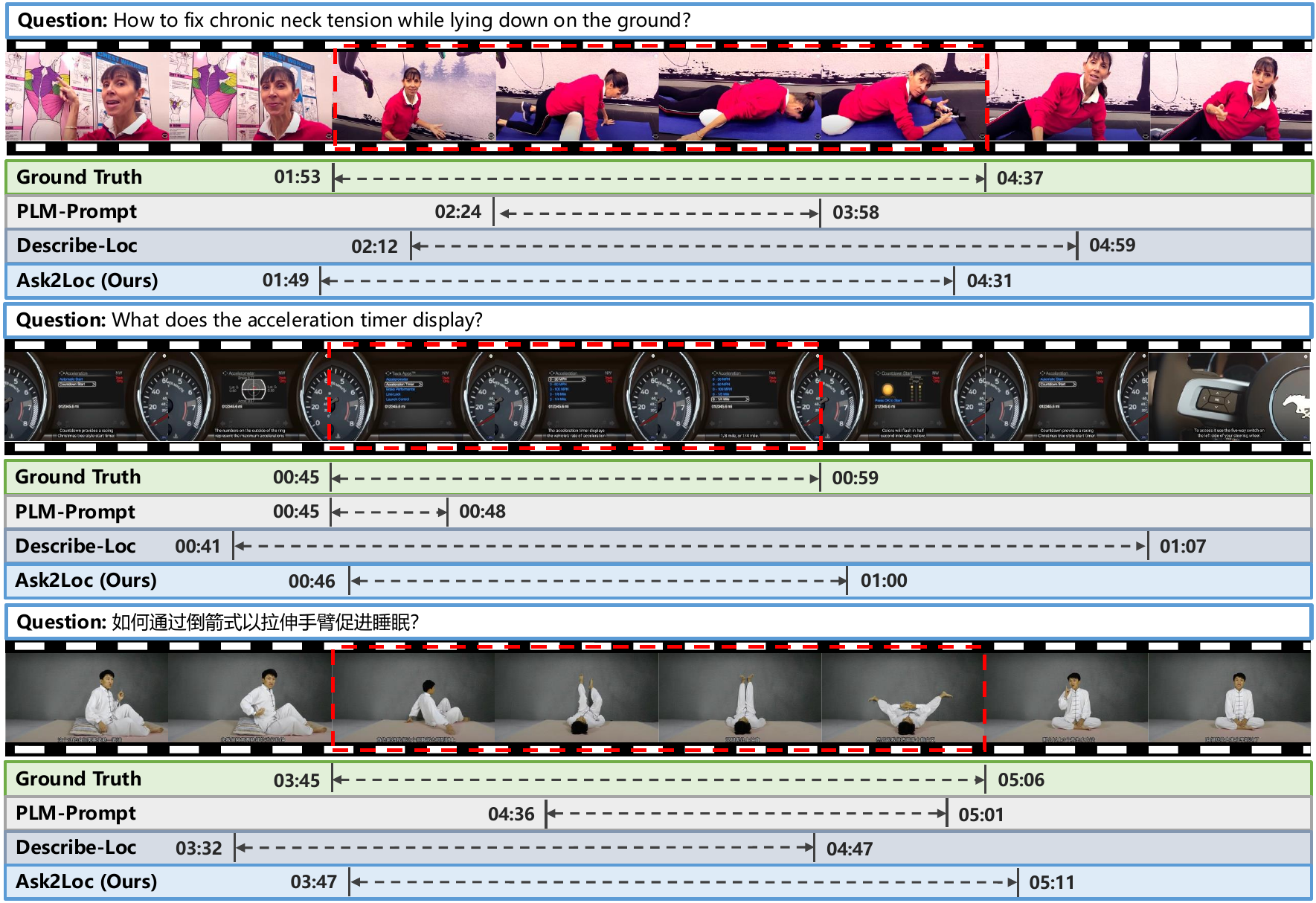}
  \caption{Three cases across multiple domains and multilingual scenarios. Our Ask2Loc demonstrates a performance improvement compared to the representative baseline methods.}
  \label{case1}
  \vspace{-10pt}
\end{figure*}

Figure \ref{case1} presents three examples from three distinct In-VAL datasets. These multi-domain and multilingual examples demonstrate that our Ask2Loc framework can acquire information through multiple interactions to support providing more accurate answer locations. In contrast, representative baseline methods tend to either offer broader locations or provide narrower spans.

\section{Interactions During Application} \label{AppendixF}
When applying Ask2Loc for a new instructional video in In-VAL, a human user can interact with the video content instead of pure language models across three modules. First, in the chatting module, the LLM generates a new detailed question based on the initial question posed by the user and the full content of the video. The user can simply responds with yes or no, and this QA continues for multiple turns to form a dialogue. Next, in the rewriting module, the user provides prompts based on his/her demands, allowing the LLM to rewrite descriptions of the question and each video segment as desired. Then, in the searching module, users can offer strategies for contextual search, such as additional prompts during the search, similarity calculation methods, top k parameters. Finally, using the additional knowledge obtained from these interactions, and the trained multimodal fusion and location detection models, Ask2Loc can provide location labels for each video segment. By extracting the answer location within the video, the system can return the corresponding visual answer clip to the user.

\end{document}